%% file: main.tex
\PassOptionsToPackage{table}{xcolor}
\documentclass[10pt,twocolumn,letterpaper]{article}

\usepackage{cvpr}
\usepackage{amsmath}
\usepackage{graphicx}
\usepackage{subcaption}
\usepackage{booktabs}
\usepackage{xcolor}
\usepackage{pifont}

\newcommand{\cmark}{\textcolor{blue}{\ding{51}}}
\newcommand{\xmark}{\textcolor{red}{\ding{55}}}

\usepackage{cuted}
\usepackage{caption}
\input{preamble}
\definecolor{cvprblue}{rgb}{0.21,0.49,0.74}
\usepackage[pagebackref,breaklinks,colorlinks,allcolors=cvprblue]{hyperref}

\title{JRDB-Pose3D: A Multi-person 3D Human Pose and Shape Estimation Dataset for Robotics}

\author{Sandika Biswas\\
Monash University\\
{\tt\small sandika.biswas@monash.edu}
\and
Kian Izadpanah\\
Sharif University\\
{\tt\small kianizadpanah@gmail.com}
\and
Hamid Rezatofighi\\
Monash University\\
{\tt\small hamid.rezatofighi@monash.edu}
}

\begin{document}
\maketitle

\pagestyle{plain}
\pagenumbering{arabic}
\input{sec/0_abstract}    
\input{sec/1_intro}

\input{sec/2_related_work}

\input{sec/3_JRDB_dataset}

\input{sec/4_Conclusion}
{
    \small
    \bibliographystyle{ieeenat_fullname}
    \bibliography{main}
}


\end{document}

%% file: sec/0_abstract.tex
\begin{strip}
    \centering
    \includegraphics[width=\textwidth]{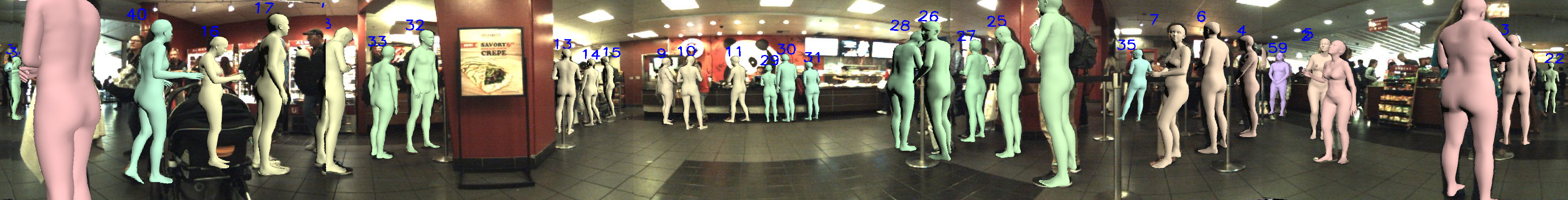}
    \includegraphics[width=\textwidth]{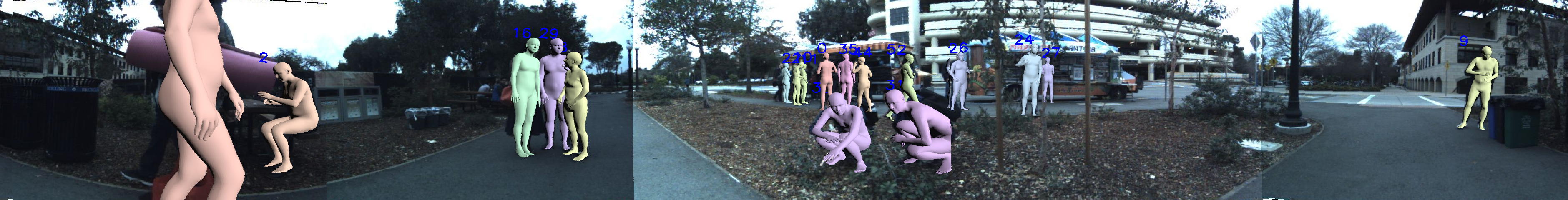}
    \captionof{figure}{Example visualization of the proposed JRDB-pose3D dataset with an indoor - \textit{'tressider-2019-04-26\_1'} (first row) and outdoor \textit{'food-trucks-2019-02-12\_0'} (second row) multi-person scene.}
    \label{fig:example}
\end{strip}
\begin{abstract}
Real-world scenes are inherently crowded. Hence, estimating 3D poses of all nearby humans, tracking their movements over time, and understanding their activities within social and environmental contexts are essential for many applications, such as autonomous driving, robot perception, robot navigation, and human–robot interactions \etc. However, most existing 3D human pose estimation datasets primarily focus on single-person scenes or are collected in controlled laboratory environments, which restricts their relevance to such real-world applications. 
To bridge this gap, we introduce JRDB-Pose3D, which captures multi-human indoor and outdoor environments from a mobile robotic platform. JRDB-Pose3D provides rich 3D human pose annotations for such complex, dynamic scenes, including SMPL-based pose annotations with consistent body-shape parameters and track IDs for each individual over time. JRDB-Pose3D contains, on average, 5–10 human poses per frame, with some scenes featuring up to 35 individuals simultaneously. The proposed dataset presents unique challenges, including frequent occlusions, truncated bodies, and out-of-frame body parts, which closely reflect real-world environments. Moreover, JRDB-pose3D inherits all the available annotations of the JRDB dataset, such as 2D pose, information about social grouping, activities, interactions, full scene semantic masks with consistent human and object level tracking, and detailed annotations for every individual, such as age, gender, and race, making it a holistic dataset for a wide range of downstream perception and human-centric understanding tasks.   
\end{abstract}

%% file: sec/1_intro.tex
\section{Introduction}
\label{sec:intro}

\begin{table*}[t]
\centering
\small
\setlength{\tabcolsep}{2pt} 
\renewcommand{\arraystretch}{1.1}
\begin{tabular}{lccccccccccc}
\toprule
Dataset & R/Sn. & In-the-wild & Env. & Tracking & SMPL & \#Camera & \#Subjects & \#Poses & Additional & Egocentric \\
        &  & & In/Out & & & S/M & & & Annot.& robot-view \\
\midrule
\rowcolor{red!15}
Shelf \cite{belagiannis20143d} & R & \xmark & In & \xmark & \xmark & 4, S &  2 & - & \xmark & \cmark \\
\rowcolor{red!15}
Campus \cite{belagiannis20143d} & R & \cmark & In & \xmark & \xmark & 4, S &  1-3 & - & \xmark &\cmark \\
\rowcolor{red!15}
4DAssociation \cite{20204DAssociation} & R & \xmark & In & \cmark & \xmark & 6, S &  2-4 & 12k & \xmark &\cmark\\
\rowcolor{red!15}
NTU-RGB+D-120 \cite{liu2019ntu} & R & \xmark & In & \xmark & \xmark & 3, S &  1-2 & 8M & ACT & \cmark\\
\rowcolor{red!15}
Panoptic Studio \cite{joo2015panoptic} & R & \cmark & In & \cmark & \xmark & 100, S &  4-6 & 1.5M & SPC & \cmark\\
\rowcolor{red!15}
KTH  \cite{kazemi2013multi}          & R & \cmark & Out & \xmark & \xmark & 1, S & 2  & 0.8k  & \xmark & \xmark\\
\rowcolor{blue!15}
JTA Dataset \cite{fabbri2018learning} & Sn. & \cmark & Out & \cmark & \xmark & 1, S &  Avg. 60 & 10M & \xmark & \cmark\\
\rowcolor{blue!15}
BEDLAM \cite{black2023bedlam}         & Sn. & \cmark & In+Out & \cmark & \cmark & 1, M  & 1-10 & 1M & Sem-M, D, HP3D & \cmark\\
\rowcolor{blue!15}
AGORA \cite{Patel:CVPR:2021}         & Sn. & \cmark & In+Out & \xmark & \cmark & 1, S  & 5-15 & 0.173M & Sem-M & \cmark\\
\rowcolor{blue!15}
WHAC \cite{yin2024whac}         & Sn. & \cmark & In+Out & \cmark & \cmark & 1, M  & 2-3 & 1.46M & \xmark & \cmark\\
3DPW \cite{von2018recovering}          & R & \cmark & In+Out & \xmark & \cmark & 1, M   & 1-2  & 75k  & \xmark & \cmark\\
RICH \cite{huang2022capturing}           & R & \xmark & Out & \cmark & \cmark & 3, S+M   & 1-2  & 85k   & DPC, Contact-labels & \cmark\\
EgoBody   \cite{zhang2022egobody}      & R & \xmark & In & \cmark & \cmark & 3-5, M   & 2  & 440k  & Eye-gaze, D &\cmark\\
WorldPose \cite{jiang2024worldpose}   & R & \cmark & Out & \cmark & \cmark & 1, M & Avg. $>10$ & 2.5M  & \xmark & \xmark\\
\midrule
JRDB-Pose3D (Ours) & R & \cmark & In+Out & \cmark & \cmark & 5, S+M & $1-36$ & 0.6M (40K) & SPC, P2D, Sem-M$^w$ & \cmark \\
 & & & & & & & & & I, AGR, ACT  & \\
\bottomrule
\end{tabular}
\caption{Comparison of 3D human pose video datasets. \textbf{R} = Real or \textbf{Sn.} = Synthetic; \textbf{In} = Indoor or \textbf{Out} = Outdoor; \textbf{S} = Static or \textbf{M} = Moving camera; \textbf{SPC} = Sparse or \textbf{DPC} = Dense 3D Point Cloud, \textbf{P2D} = 2D pose, \textbf{Sem-M} = Segmentation mask for only human, \textbf{Sem-M$^w$} = Segmentation mask for whole scene, \textbf{D} = Depth Map, \textbf{HP3D} = 3D Hand Pose, \textbf{I} = Human-human or human-scene interactions, \textbf{AGR} = Age-Gender-Race, \textbf{ACT} = Activities. Datasets marked in red provide 3D skeleton annotations for human pose. Datasets marked in blue do not contain coordinated human poses within a frame and meaningful interactions with the scene.}
\label{tab:dataset_compare}
\end{table*}

Real-world scenes typically involve multiple humans, making multi-person 3D pose estimation and tracking in unconstrained, in-the-wild environments a fundamental research problem across a wide range of applications, such as autonomous driving, robotic perception and navigation, and human–robot interaction. Correct interpretation of such scenes requires a comprehensive understanding of diverse scene elements, including crowd behavior, global human trajectories, social interactions, social groupings, group-level activities, and human interactions with the environment and other individuals. These tasks are challenging, especially in crowded scenes, due to large variations in human motion and frequent occlusions arising from interactions. 

3D Human pose and shape estimation has shown significant advancements in computer vision research. However, both existing methods and widely used benchmark datasets predominantly focus on single human activities in the scene \cite{h36m_pami, Sigal2006HumanEva, Sigal2010HumanEvaII}. In recent years, a growing body of work has proposed multi-human pose estimation datasets. Some of these focus only on the interactions of two people \cite{liu2019ntu,zhang2022egobody,Fieraru_2020_CVPR,fieraru2023_interactions_arxiv,hassan2019resolving,von2018recovering,huang2022capturing}, while others provide only 3D skeleton annotations \cite{kazemi2013multi,belagiannis20143d,fabbri2018learning,mehta2018single,20204DAssociation,park2023towards,joo2015panoptic}. A few methods propose building synthetic datasets \cite{black2023bedlam,Patel:CVPR:2021,yin2024whac} with full body annotation including both shape and poses. More recent approaches, Crowd3D \cite{wen2023crowd3d} captures a large-scale multi-person scene, but are limited to an image dataset. WorldPose \cite{jiang2024worldpose} introduces a multi-human video dataset, but it is constrained by the camera views and scene diversity. Moreover, this dataset captures a specific type of scene, such as a sports scene, and poses are far from the camera and captured mainly from the top view, limiting its applicability to generic crowded-scene pose estimation and long-term multi-person tracking in everyday environments.

In this paper, we introduce JRDB-Pose3D, a real-world dataset for 3D multi-human pose and shape estimation built upon the JRDB dataset \cite{martin2021jrdb}. JRDB \cite{martin2021jrdb} is a large-scale dataset captured from the JackRabbot mobile robot platform while it navigates through a university campus, recording crowded real-world scenes containing multiple humans. The robot is equipped with five pairs of stereo cameras, providing a 360° panoramic view of its surroundings. This dataset comprises 54 sequences collected across both indoor and outdoor settings. As the robot moves through these scenes, it captures a wide variety of human poses, ranging from distant to close-range views, partial and heavily occluded poses, human–human and human-scene interactions. JRDB-Pose3D provides SMPL-based 3D pose and shape annotation on the JRDB dataset while inheriting its complementary annotations of crowd-level social dynamics, social grouping, group activities \cite{ehsanpour2022jrdb}, dense semantic segmentation and long-term tracking of all the scene elements \cite{le2024jrdbpanotrack}, detailed human–human and human–environment interactions \cite{jahangard2024jrdbsocial}, as well as long-term global human trajectories across the scene. With these rich additional annotations, JRDB-pose3D provides an ideal testbed for a wide range of downstream tasks, including multi-person pose estimation, interaction-aware pose estimation, and pose-aware activity recognition in a real-world crowd environment. 
Our contributions are:
\begin{itemize}
    \item We introduce JRDB-Pose3D, a large-scale multi-person 3D human pose-shape (HPS) estimation dataset with consistent per subject tracking IDs and per-keypoint occlusion labels.
    \item We evaluate state-of-the-art multi-person pose estimation, pose tracking, and action-aware pose prediction methods on the JRDB-Pose3D dataset to assess the strengths and weaknesses of the dataset.
\end{itemize}

%% file: sec/2_related_work.tex
\section{Related Datasets} 
In this section, we discuss about the most commonly used datasets for 3D human pose estimation - single \cite{h36m_pami,Sigal2006HumanEva,Sigal2010HumanEvaII,mehta2017monocular,ghorbani2021movi} or multiple human pose estimations \cite{belagiannis20143d,20204DAssociation,liu2019ntu,joo2015panoptic,fabbri2018learning,kazemi2013multi,black2023bedlam,Patel:CVPR:2021,yin2024whac,von2018recovering,hassan2019resolving,zhang2022egobody,huang2022capturing,jiang2024worldpose}, and tracking.  

\subsection{Single human pose estimation} 3D human pose estimation has been widely studied in the computer vision community. Early benchmark datasets primarily focused on scenes containing a single human subject. For instance, Human3.6M \cite{h36m_pami}, a multi-view dataset, captures sequences of basic daily-life activities such as walking, sitting, and eating. HumanEva-I \cite{Sigal2006HumanEva} includes videos of a more diverse set of activities, e.g., walking, jogging, boxing, throwing, and catching, while HumanEva-II \cite{Sigal2010HumanEvaII} extends this setup by capturing combinations of these activities. MPI-INF-3DHP \cite{,mehta2017monocular}, Movi \cite{ghorbani2021movi} gradually move towards capturing in-the-wild poses of a single human in indoor/ outdoor environments. Although these datasets are useful for building robust single-human-pose estimation models, their applicability to complex, crowded scenes is limited because most of these datasets were collected in controlled laboratory environments and do not capture the dynamics of natural environments. 
 In contrast, JRDB-Pose3D provides 3D human pose annotations for a robot navigation dataset capturing real-world in-the-wild scenes, with a multi-person environment.

\subsection{Multiple Human Pose Estimation}
In recent years, a growing body of research has focused on multi-human pose estimation, as it more closely reflects real-world scenes. Existing datasets in this category can be broadly categorized based on the level of detail provided in their annotations.

\subsubsection{3D Skeleton-based Annotations}
The Shelf dataset~\cite{belagiannis20143d} is one of the earliest benchmarks for capturing interactions among multiple humans using a multi-view camera setup. It contains scenes with two interacting subjects recorded in an indoor environment. The Campus dataset~\cite{belagiannis20143d}, proposed by the same authors, extends this setup to outdoor environments. However, both datasets are relatively small and exhibit limited variability in terms of scene complexity and interaction types.

The KTH dataset~\cite{kazemi2013multi} focuses primarily on sports-related scenes and contains approximately 800 annotated poses. The 4D Association dataset~\cite{20204DAssociation} captures interactions among up to five individuals per scene and provides five sequences of approximately 20 seconds each, yielding roughly 12K annotated poses. In contrast, the JTA dataset~\cite{fabbri2018learning} is significantly larger, comprising nearly 10 million poses; however, it is entirely synthetic, which limits its domain transferability.

All of these datasets provide 3D joint annotations but lack explicit modeling of human body shape. Detailed body shape information is essential for accurately reasoning about occlusions, physical interactions, and spatial relationships between individuals and their environments. Without shape-aware representations, many downstream tasks, such as interaction understanding, contact reasoning, collision prediction, and scene-level physical consistency, become fundamentally constrained.

\subsubsection{3D Human Pose and Shape (HPS) Annotations}
With the advent of parametric human body models such as SMPL and SMPL-X, several datasets now provide annotations that jointly encode human pose and body shape. These representations enable more complete reasoning about human geometry compared to skeletal-only annotations. These datasets can be broadly categorized as follows, 

\noindent
\textbf{Synthetic datasets:} Due to difficulty in capturing multi-human scenes with ground-truth 3D poses, many research has proposed generating synthetic datasets. AGORA~\cite{Patel:CVPR:2021} fits SMPL-X models to commercially available high-quality 3D scans of humans and renders them to generate a large-scale multi-human synthetic dataset. Each image contains between 5 and 15 individuals. However, AGORA is an image-based dataset that lacks temporal continuity, making it unsuitable for human tracking or motion forecasting. BEDLAM~\cite{black2023bedlam} proposes another large-scale synthetic dataset, in which human poses sampled from the AMASS dataset are placed within diverse 3D environments. 

Although both datasets contain multiple humans per scene, the poses are not coordinated across individuals; therefore, they do not capture any meaningful inter-person interactions. Although synthetic datasets are valuable for building robust HPS estimation models, they fail to ensure domain transferability with robust pose estimations under real-world complexities. In contrast, JRDB-pose3D captures real-world robot navigation scenes.
 \\
 \noindent
\textbf{Real datasets:} Some research has introduced real-world datasets containing multiple humans; however, most are limited to scenes involving interactions between only two individuals. 3DPW \cite{von2018recovering} captures outdoor multi-human scenes but provides annotations for at most two persons per frame. RICH \cite{huang2022capturing} primarily focuses on human–scene interactions in both indoor and outdoor environments, offering detailed annotations of human body vertices that come into contact with the scene during interaction. While a small subset of its sequences includes interactions between two people, multi-person interactions are not its primary focus. EgoBody \cite{zhang2022egobody} captures social interactions in 3D scenes involving two interacting subjects, providing both egocentric views and third-person views of the scene.
In contrast, WorldPose \cite{jiang2024worldpose} is the first video dataset to capture large-scale scenes with a high density of people, averaging more than 10 individuals per frame. It annotates a 50-minute FIFA World Cup video, resulting in approximately 2.5 million 3D poses. However, the dataset is recorded exclusively from a top-down camera viewpoint, which limits its applicability to tasks such as robot navigation, where perception from ground-level or robot-mounted cameras is essential. JRDB-pose3D also captures large-scale human poses per frame from robot view, perfect for robot perception applications.
\\
\noindent
\textbf{Datasets with Additional Annotations:} The majority of the above-mentioned datasets only provide annotations for human pose. Among the synthetic datasets, BEDLAM and AGORA provide additional annotations for semantic segmentation masks of humans, while BEDLAM and Ego-body provide information for hand pose annotations and tracking. However, none of these datasets give information about human-human or human-scene interactions. On the other hand, JRDB-pose3D inherits all the available annotations of JRDB dataset \cite{martin2021jrdb} such as, 2D pose annotation \cite{vendrow2023jrdb}, group level activities, interactions \cite{ehsanpour2022jrdb}, full scene semantic masks with consistent human and object level tracking \cite{le2024jrdbpanotrack}, detailed annotations for every individual's activities, their age, gender, race \cite{jahangard2024jrdbsocial} etc. Together, these annotations enable a comprehensive understanding of the real-world scenes.

%% file: sec/3_JRDB_dataset.tex
\section{JRDB-Pose3D Dataset}


\subsection{Dataset overview}
We construct JRDB-Pose3D by annotating all video sequences from the JRDB dataset \cite{martin2021jrdb}. JRDB contains 54 sequences captured in real-world, multi-human environments as a mobile robot navigates through a university campus. The data set is collected using the JackRabbot robotic platform, which is equipped with stereo camera pairs arranged in two horizontal rows, with each row consisting of five RGB cameras at a resolution of $752 \times 480$. Together, these five cameras of each row provide a full panoramic view, forming two stereo panoramic views (each with resolution $3760 \times 480$) of the scene. The dataset contains a total of 2,88,435 frames with 636K 2D human poses and 5022 individual tracks. JRDB-Pose3D annotates all humans that already have 2D skeleton pose \cite{vendrow2023jrdb} and 3D bounding box annotation \cite{martin2021jrdb}, while every 15th frame poses are manually checked and corrected if necessary. JRDB-Pose3D contains SMPL pose, consistent body shape for individual tracks, global positioning of the tracks in the 3D lidar point cloud, consistent track ID, and gender information.
\subsection{Annotation} The dataset is annotated using a multi-stage pipeline consisting of: (1) pose initialization with pretrained state-of-the-art models, (2) localization of individual poses in the global 3D scene, (3) consistent shape annotation of individual tracks across frames, (4) refinement of local 3D poses via optimization, and (5) manual inspection and correction of the final poses. \\
\noindent
\textbf{Pose initialization:} In the first stage, we initialize 3D SMPL poses using the pretrained CameraHMR model \cite{prianka3dv2025}, which is specifically trained for multi-human scenes. CameraHMR predicts the camera Field-of-View (FoV) directly from the input image and uses this estimated FoV during pose optimization, which enables more accurate 3D pose prediction. As a result, it produces reliable pose estimates in crowded, multi-person scenarios. However, CameraHMR operates on single images and therefore does not enforce identity tracking across a sequence. To address this limitation, we leverage the JRDB dataset’s 2D bounding boxes with consistent track IDs as input, enabling coherent 3D SMPL pose initialization $\mathbf{P}^{init}_i(\theta, t, \beta)$ for every individual track $i$ over time.\\
\noindent
\textbf{Localization in global 3D scene:} In the second stage, we place the predicted 3D poses into the global 3D scene. While CameraHMR predicts 3D poses using an estimated field-of-view (FoV), JRDB provides accurate camera intrinsics. We therefore align the predicted 3D poses to the JRDB scenes using the dataset-provided camera parameters. Specifically, we estimate a rigid transformation that aligns the initial 3D skeleton joints predicted by CameraHMR with the corresponding 2D keypoints
annotated in JRDB. Given the regressed 3D joint locations $X^{init}_{3D} \in \mathbb{R}^{17 \times 3}$ from initial SMPL pose estimation $\mathbf{P}^{init}_i(\theta, t, \beta)$ of previous stage, 2D pose annotations $x_{2D} \in \mathbb{R}^{17 \times 2}$, and camera intrinsics $K$ of JRDB, we estimate the rigid transformation:

\begin{equation}
    R, t = PnP(\mathbf{X}^{init}_{3D}, \mathbf{x}_{2D}, K)
\end{equation}
which maps SMPL coordinates into the camera frame.
The final global orientation and position of each pose are obtained by composing the estimated transformation with the initial estimates $R^{init}$ and $t^{init}$: 
\begin{equation}
    R_{global} = R R^{init}
\end{equation}
\begin{equation}
    t_{global} = t + Rt^{init}
\end{equation}
\\
\noindent
\textbf{Consistent shape annotation:} Since CameraHMR performs per-frame estimation, it does not enforce shape consistency and may produce varying body shapes for the same individual across frames. To obtain a consistent and accurate shape representation for each person in the scene, we first visually inspect the per-frame shape predictions for each track and select the frame with the best-fitting shape as the representative shape for every individual. When the selected shape is still not satisfactory, we further refine it using an annotation tool to manually adjust the shape parameters, ensuring accurate shape representation for every individual. \\
\noindent
\textbf{Refinement of local 3D pose:} After finalizing the track shape, global orientation, and camera translation, we refine the local poses. We optimize the SMPL pose initialization of the first stage by minimizing the distance of projected 3D joints with respect to the 2D pose annotation of the JRDB-pose \cite{vendrow2023jrdb}. 
\begin{equation}
    \mathbb{E}_{data} = ||\Pi(\mathbf{X}_{3D}) - \mathbf{x}_{2D}||^2_2
\end{equation}
We also regularize the poses to ensure temporal consistencies between frames, 
\begin{equation}
    \mathbb{E}_{smooth} = \sum_t ||(\mathbf{X}^{t+1}_{3D} - \mathbf{X}^{t}_{3D}) - (\mathbf{X}^{t}_{3D} - \mathbf{X}^{t-1}_{3D})||^2_2
\end{equation}
This term penalizes second-order temporal differences, encouraging smooth motion trajectories while preserving dynamics.
\\
\noindent
\textbf{Manual inspection and correction:} To ensure high-quality and consistent annotations, multiple trained annotators are assigned to review the generated poses and perform corrections when necessary using an annotation tool, following a shared annotation protocol. 

During this inspection, the annotators mainly verify the correctness and plausibility of limb orientation and its consistency with the available 2D pose annotations; temporal consistency of body poses or plausibility in the context of interactions with the environment or other humans in the scene, with more attention for the challenging cases such as heavy occlusions, out-of-frame body poses.

This manual verification stage plays a crucial role in identifying and correcting the failure cases generated by automated pose estimation and helps enforce temporal coherence, with high-quality, reliable 3D pose annotations suitable for benchmarking multi-person 3D pose estimation and downstream perception tasks. 

\begin{figure}[t]
    \centering
    \begin{subfigure}{0.5\linewidth}
        \centering
        \includegraphics[width=\linewidth]{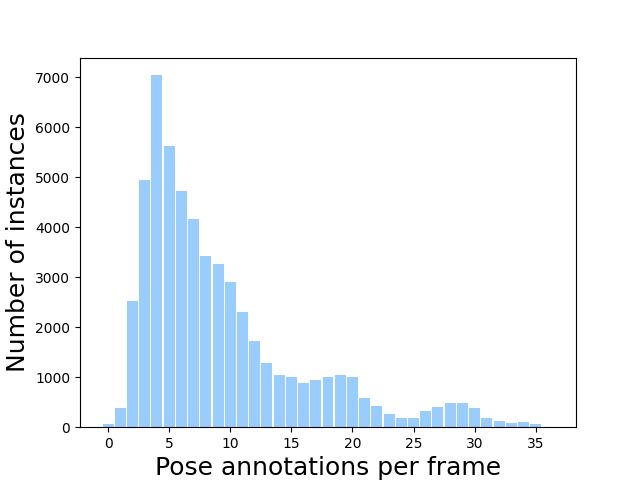}
        \caption{}
        \label{fig:ours}
    \end{subfigure}
    \begin{subfigure}{0.49\linewidth}
        \centering
        \includegraphics[width=\linewidth]{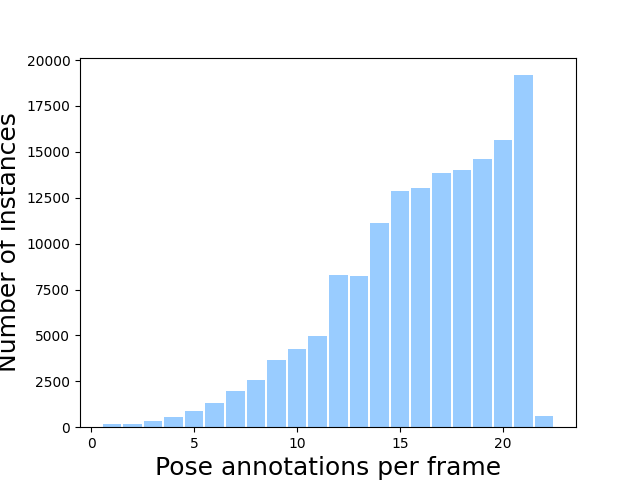}
        \caption{}
        \label{fig:wp}
    \end{subfigure}
    \caption{Statistics of the per-frame number of poses for (a) Ours and (b) WorldPose dataset.}
    \label{fig:pose_stats}
\end{figure}

\noindent
\section{Uniqueness of JRDB-Pose3D}
In contrast to the existing state-of-the-art multi-person datasets, JRDB-Pose3D uniquely reflects real-world robotic perception scenarios in a crowded environment. Figure \ref{fig:pose_stats} compares the number of poses per-frame in JRDB-Pose3D with WorldPose, a dataset that provides a setup closely related to ours. JRDB-Pose3D demonstrates a long-tailed distribution, with frames containing up to 35 poses, highlighting the increased crowd density and complexity present in our dataset. 

\begin{figure}[ht]
    \centering
    \begin{subfigure}{0.5\linewidth}
        \centering
        \includegraphics[width=\linewidth]{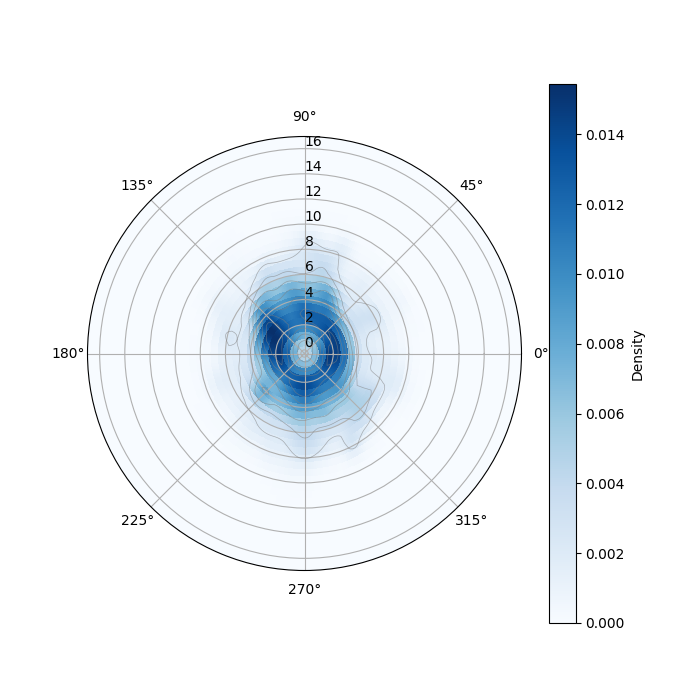}
        \caption{}
        \label{fig:ours_kde}
    \end{subfigure}
    \begin{subfigure}{0.49\linewidth}
        \centering
        \includegraphics[width=\linewidth]{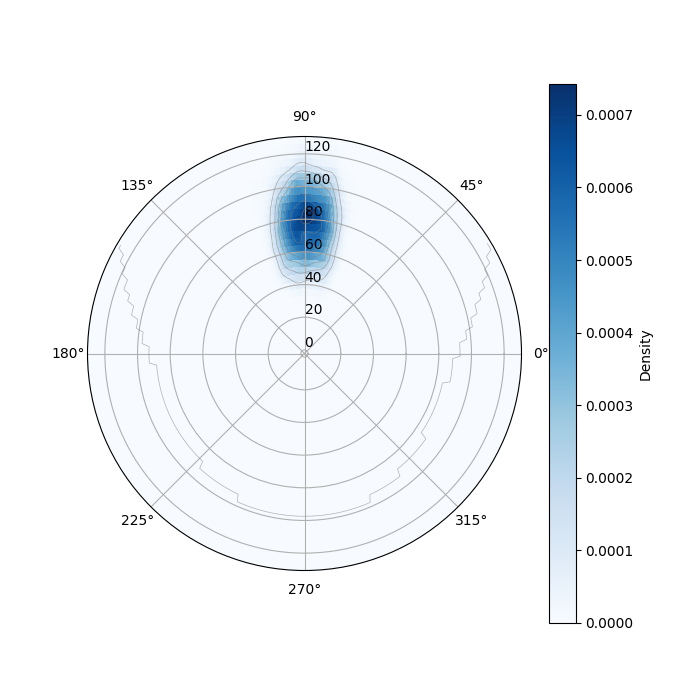}
        \caption{}
        \label{fig:wp_kde}
    \end{subfigure}
    \caption{Kernel Density Estimates (KDEs) of the spatial distribution of people around the robot, i.e., distance from the camera and angular positions, across (a) Ours and (b) WorldPose dataset.}
    \label{fig:kdes}
\end{figure}

Figure \ref{fig:kdes} shows a comparison of the angle and distance distribution of the poses from the camera. JRDB-Pose3D offers a panoramic view of the scene with human poses distributed across a $360^{\circ}$ span within an average 5-10m distance from the camera. The WorldPose dataset is captured under play ground environment from a distant camera, hence majorly the poses are very far from the camera, with an average distance of around 80m, and captured from a particular angle, exhibiting a limited angular distribution of the poses around the scene. The polar density plot reveals a strong concentration of poses directly in front of the camera, with the highest density centered around $90^{\circ}$.

\begin{figure}[ht]
    \centering
    \begin{subfigure}{0.5\linewidth}
        \centering
        \includegraphics[width=\linewidth]{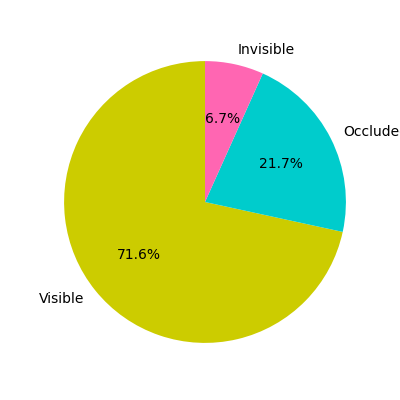}
        \caption{}
        \label{fig:occ_vs_vis}
    \end{subfigure}
    \begin{subfigure}{0.49\linewidth}
        \centering
        \includegraphics[width=\linewidth]{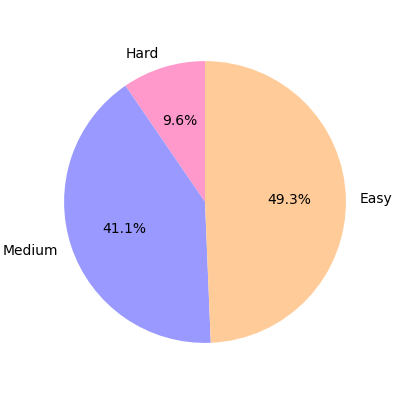}
        \caption{}
        \label{fig:hardness}
    \end{subfigure}
    \caption{(a) Statistics of occlusion and out-of-frame poses in the JRDB-Pose3D dataset in terms of the number of occluded, invisible, and visible 2D joints. (b) Statistics of poses in the JRDB-Pose3D dataset according to the difficulty of 3D pose recovery, classified as easy, medium, and hard.}
    \label{fig:occ_stats}
\end{figure}

The JRDB dataset is captured by a human-sized mobile robot navigating through a university campus. As a result, JRDB-Pose3D frequently contains crowded indoor and outdoor environments, leading to human poses with prolonged partial and full occlusions caused by other humans and scene objects. In addition, while navigating or interacting, humans often appear at very close proximity to the robot, resulting in a large number of out-of-frame poses.
These factors make the dataset particularly challenging and realistic for both multi-person tracking and 3D human pose and shape (HPS) estimation tasks.
Figure \ref{fig:occ_vs_vis} presents statistics on the number of visible, occluded, and out-of-frame (invisible) 2D joints per frame, highlighting the inherent challenges of the dataset. Furthermore, Figure \ref{fig:hardness} summarizes the distribution of poses as easy, medium, or hard for 3D pose recovery, categorized based on the number of occluded and out-of-frame joints per pose.

%% file: sec/4_Conclusion.tex
\section{Conclusion}

This paper introduces JRDB-Pose3D, an in-the-wild multi-person 3D human pose and shape estimation dataset designed to capture the complexities of perception tasks in real-world conditions. Collected using a mobile robotic platform navigating through indoor and outdoor environments, JRDB-Pose3D departs from controlled laboratory settings and instead emphasizes crowded, dynamic environments representative of practical real-world scenes.
JRDB-Pose3D provides temporally consistent SMPL-based 3D pose and shape annotations with identity tracking, supporting scenarios with high densities of interacting individuals. The dataset presents significant challenges, caused by severe occlusions, out-of-frame body parts, and complex human–human and human–object interactions, which makes it well-suited for benchmarking robust multi-person 3D pose estimation, tracking, and pose forecasting methods in unconstrained or in-the-wild environments. Moreover, with additional multimodal annotations from the JRDB dataset, including 2D pose, activity labels, social grouping, interaction annotations, and full-scene semantic segmentation, JRDB-Pose3D makes it a perfect benchmark for holistic human-centric scene understanding.
We believe JRDB-Pose3D will serve as a valuable resource for advancing research on socially aware downstream tasks, particularly those requiring an understanding of multi-human interactions in complex real-world environments.